# Deep-LK for Efficient Adaptive Object Tracking

Chaoyang Wang, Hamed Kiani Galoogahi, Chen-Hsuan Lin and Simon Lucey
Carnegie Mellon University

## Abstract

*In this paper we present a new approach for efficient regression based object tracking which we refer to as Deep-LK. Our approach is closely related to the Generic Object Tracking Using Regression Networks (GOTURN) framework of Held et al. [16]. We make the following contributions. First, we demonstrate that there is a theoretical relationship between siamese regression networks like GOTURN and the classical Inverse-Compositional Lucas & Kanade (IC-LK) algorithm. Further, we demonstrate that unlike GOTURN IC-LK adapts its regressor to the appearance of the currently tracked frame. We argue that this missing property in GOTURN can be attributed to its poor performance on unseen objects and/or viewpoints. Second, we propose a novel framework for object tracking - which we refer to as Deep-LK - that is inspired by the IC-LK framework. Finally, we show impressive results demonstrating that Deep-LK substantially outperforms GOTURN. Additionally, we demonstrate comparable tracking performance to current state of the art deep-trackers whilst being an order of magnitude (i.e. 100 FPS) computationally efficient.*

## 1. Introduction

Regression based strategies to object tracking have long held appeal from a computational standpoint. Specifically, they apply a computationally efficient regression function to the current source image to predict the geometric distortion between frames. As noted by Held et al. [16] most the state of the art trackers in vision are based on a classification-based approach, classifying many image patches to find the target object. Notable examples in this regard are correlation filter methods [13, 17, 19, 6, 11], and more recent extensions based on deep learning [7, 30, 37] can be also considered as adhering to this classification-based paradigm to tracking.

Recently, Held et al. [16] proposed their Generic Object Tracking Using Regression Networks (GOTURN) framework for object tracking. As it is a regression based approach, it can be made to operate extremely efficiently - 100 Frames Per Second (FPS) on a high-end GPU. This is in stark contrast to the latest classification-based deep networks for tracking - e.g. MDNet [30], ECO [11] report tracking times of less than 10 FPS using a GPU. Classification-based methods to object tracking that do not rely on deep learning - such as correlation filters - can compete with GOTURN in terms of computational efficiency. However, they often suffer from poor performance as they do not take advantage of the large number of videos that are readily available to improve their performance. Finally, regression based strategies hold the promise of being able to efficiently track objects with more sophisticated geometric deformations than just translation (e.g. affine, homographies, thin-plate spline, etc.) - something that is not computationally feasible with classification-based approaches.

Although showing much promise, GOTURN has a fundamental drawback. As noted by Held et al. the approach performs admirably on object's that have similar appearance to those seen in training. However, when attempting to apply GOTURN to object's or viewpoints that have not been seen in training the approach can dramatically fail. This type of poor generalization is not seen in classification-based deep networks such as FCSN etc [7]. This poor result at first glance is surprising as GOTURN employs a siamese network which predicts the relative geometric deformation between frames. This network architecture, along with the employment of tracking datasets with large amounts of object variation, should in principle overcome this generalization issue. It is this discrepancy between theory and practice which is at the heart of our paper.

In this paper we attempt to explain this discrepancy by looking back to a classical algorithm in computer vision literature - namely the Lucas & Kanade (LK) algorithm [25]. Specifically, we demonstrate that Inverse-Compositional LK (IC-LK) [4] can be re-interpreted as a siamese regression network that shares many similar properties to GOTURN. We note, however, an important difference that drives the central thesis of our paper. Specifically, within the IC-LK algorithm the regression function adapts to the appearance of the previously tracked frame. In GOTURN, their regression function is "frozen" - partially explaining why the approach performs so poorly on previously unseen objects and/or viewpoints. Based on this insight we propose a new algorithm for efficient object tracking - which we refer to as DeepLK - that enjoys the same computational advantages of GOTURN without the poor generalization.

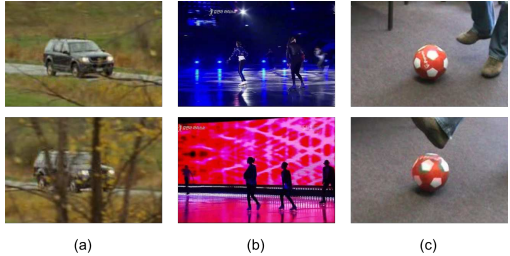

Figure 1. Three typical types of variations which LK is sensitive to: (a) occlusion; (b) illumination and appearance changes; (c) unmodeled geometric transformation, such as out-of-plane rotation of the ball.

When evaluated upon VOT2014, our DeepLK method achieves very competitive results to the state of the art classification-based deep methods (e.g. FCSN), and significantly outperforms GOTURN. Further, we demonstrate the superior generalization properites of DeepLK in comparison to GOTURN. We compare with state-of-the-art trackers on 50 challenging higher frame rate videos (240 FPS) from the NfS [1] tracking dataset. For these higher frame rate videos, our proposed approach is on par with state-of-the-art classification-based deep trackers while enjoying an order of magnitude faster (100 FPS) computational efficiency.

**Notations.** We define our notations throughout as follows: lowercase boldface symbols (*e.g.* $\mathbf{x}$) denote vectors, uppercase boldface symbols (*e.g.* $\mathbf{W}$) denote matrices, and uppercase calligraphic symbols (*e.g.* $\mathcal{I}$) denote images, and we use the notations $\mathcal{I}(\mathbf{x}) : \mathbb{R}^2 \to \mathbb{R}^K$ to indicate sampling of the $K$-channel image representation at subpixel location $= [x, y]^\top$.

## 2. The Inverse Compositional LK Algorithm

The core assumption of LK is that the relationship of the image appearance has an approximated linear relationship with the geometric displacements. Given a predefined geometric warp function parametrized by the warp parameters $\mathbf{p}$, this relationship can be written as

$$\mathcal{I}(\Delta \mathbf{p}) \approx \mathcal{I}(\mathbf{0}) + \frac{\partial \mathcal{I}(\mathbf{0})}{\partial \mathbf{p}} \Delta \mathbf{p} , \quad (1)$$

which is the first-order Taylor expansion evaluated at the identity warp $\mathbf{p} = \mathbf{0}$.

Given a source image $\mathcal{I}$ and a template image $\mathcal{T}$ to align against, the Inverse Compositional (IC) LK algorithm attempts to find the geometric warp update $\Delta \mathbf{p}$ on $\mathcal{T}$ that could in turn be inversely composed to $\mathcal{I}$ by minimizing the sum of squared differences (SSD) objective

$$\min_{\Delta \mathbf{p}} \|\mathcal{I}(\mathbf{p}) - \mathcal{T}(\Delta \mathbf{p})\|_2^2 . \quad (2)$$

IC-LK further utilizes Equation 1 to linearize Equation 2 as

$$\min_{\Delta \mathbf{p}} \|\mathcal{I}(\mathbf{p}) - \mathcal{T}(\mathbf{0}) - \frac{\partial \mathcal{T}(\mathbf{0})}{\partial \mathbf{p}} \Delta \mathbf{p}\|_2^2 . \quad (3)$$

Solving for Equation 3, we get the IC-LK update equation:

$$\Delta \mathbf{p} = \mathbf{W}^\dagger [\mathcal{I}(\mathbf{p}) - \mathcal{T}(\mathbf{0})]. \quad (4)$$

where † is the Moore-Penrose pseudo-inverse operator. $\mathbf{W}$ is the so-called template image Jacobian, factorized as

$$\mathbf{W} = \frac{\partial \mathcal{T}(\mathbf{0})}{\partial \mathbf{p}} = \nabla \mathcal{T}(\mathbf{0}) \frac{\partial \mathcal{W}(\mathbf{x}; \mathbf{0})}{\partial \mathbf{p}^\top} , \quad (5)$$

where $\nabla \mathcal{T}(\mathbf{0})$ are the image gradients that are solely a function of $\mathcal{T}(\mathbf{0})$, and $\frac{\partial \mathcal{W}(\mathbf{x}; \mathbf{0})}{\partial \mathbf{p}^\top}$ is the predefined warp Jacobian. The warp parameter $\mathbf{p}$ are updated by:

$$\mathbf{p} \leftarrow \mathbf{p} \circ^{-1} \Delta \mathbf{p}, \quad (6)$$

where we use the notation $\circ^{-1}$ to denote the inverse compositional warp functions parametrized by $\mathbf{p}$, and $\Delta \mathbf{p}$.

Since the true relationship between appearance and geometric deformation is not linear in most cases, Equations 4 and 6 need to be applied iteratively until convergence.

The advantage of the IC form of LK over the original lies in its efficiency. Specifically, $\mathbf{W}^\dagger$ is evaluated on the template image $\mathcal{T}$ at the identity warp $\mathbf{p} = \mathbf{0}$; therefore, $\mathbf{W}^\dagger$ remains constant throughout as long as the template image remains static. We refer the readers to [4] for a more detailed treatment.

Recently, there has been a number of works that aims to increase the robustness of the classical LK framework by minimizing feature differences extracted from the images [2, 3]. Following the same aforementioned derivation, the geometric warp update $\Delta \mathbf{p}$ in Equation 4 becomes

$$\Delta \mathbf{p} = \mathbf{W}^\dagger [\phi(\mathcal{I}(\mathbf{p})) - \phi(\mathcal{T}(\mathbf{0}))] , \quad (7)$$

where $\phi(\cdot)$ is a feature extraction function; correspondingly, $\mathbf{W} = \frac{\partial \phi(\mathcal{T}(\mathbf{0}))}{\partial \mathbf{p}}$ is also solely a function of $\phi(\mathcal{T}(\mathbf{0}))$. Similar to the original IC-LK, $\mathbf{W}^\dagger$ is fixed throughout since $\phi(\mathcal{T}(\mathbf{0}))$ only needs to be evaluated once at the identity warp.

### 2.1. Cascaded Linear Regression

Equation 7 can be written in a more generic form of linear regression as

$$\Delta \mathbf{p} = \mathbf{R} \cdot \phi(\mathcal{I}(\mathbf{p})) + \mathbf{b}, \quad (8)$$

where $\mathbf{R} = \mathbf{W}^{\dagger}$ and $\mathbf{b} = -\mathbf{W}^{\dagger} \cdot \phi(\mathcal{T}(\mathbf{0}))$ are the regression matrix and the bias term respectively. Therefore, the IC-LK algorithm belongs to the class of cascaded linear regression methods, as the warp update is iteratively solved for and applied until convergence.

Cascaded regression has been widely used in pose estimation and face alignment. Compared to directly solving the problem with a single sophisticated model (*e.g.* a deep fully connected neural network), cascaded regression can usually achieve comparable results with a sequence of much simpler models [14, 40, 35, 8]. This is desirable to visual tracking, as in many cases, simpler models are computationally more efficient and thus offer faster speed.

### 2.2. Connection to Siamese Regression Networks

Siamese regression networks for tracking predict the geometric warp from concatenated features of the source and template images together. This can be mathematically expressed as

$$\Delta \mathbf{p} = f(\begin{bmatrix} \phi(\mathcal{I}(\mathbf{p})) \\ \phi(\mathcal{T}(\mathbf{0})) \end{bmatrix}) \ , \quad (9)$$

where $f(\cdot)$ denotes a nonlinear prediction model . In the case of GOTURN [16], $\phi(\cdot)$ is the convolutional features and $f(\cdot)$ is a multi-layer perceptron trained through back-propagation.

We can also write the IC-LK formulation in Equation 7 in the form of Equation 9 as

$$\Delta \mathbf{p} = \begin{bmatrix} \mathbf{W}^{\dagger} & -\mathbf{W}^{\dagger} \end{bmatrix} \begin{bmatrix} \phi(\mathcal{I}(\mathbf{p})) \\ \phi(\mathcal{T}(\mathbf{0})) \end{bmatrix} . \quad (10)$$

This insight elegantly links the IC-LK algorithm with GOTURN: while GOTURN models the non-linear prediction function $f$ with a heavily-parametrized multi-layer perceptron, IC-LK models $f$ through cascaded linear regression. In addition, $\mathbf{W}^{\dagger}$ is a sole function of $\mathcal{T}(\mathbf{0})$ and contains no learnable parameters.

In comparison to GOTURN whose weights are offline learned and kept static during testing, the fully connected layer of IC-LK is formed directly from the template image. In some sense, this is analogous to one-shot learning, since the weights are "learned" from one single example $\mathcal{T}(\mathbf{0})$. This is a very desirable property for tracking, as it allows the tracker to cheaply adapt its parameters in real-time, and as a result, generalize better to unseen objects.

### 2.3. Limitations

Although having these desired properties, prior works on LK have the following limitations (illustrated in Fig. 1): they are (a) sensitive to occlusion; (b) sensitive to appearance variations; (c) intolerant to geometric transformations which are not modeled by the warp Jacobian(*e.g.* out of plane rotation and pose changes of articulated objects).

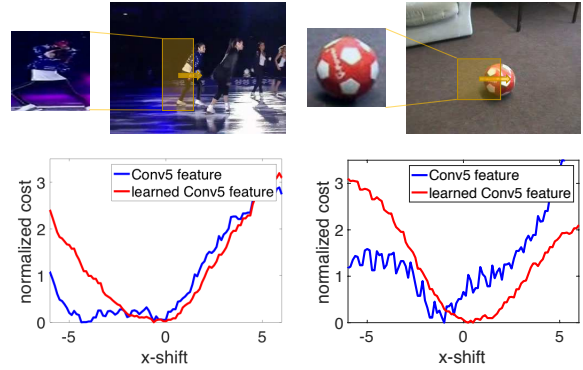

Figure 2. Cost curves of vanilla AlexNet's conv5 feature (blue) and Deep-LK's learned feature (red) along horizontal shifts, computed in terms of SSD. The curves of learned features exhibit less local minima and encourages smoother convergence. This characteristic is required by the LK framework.

All these limitations severely restrict the application of LK framework in real world settings.

For the past three decades, there have been numerous attempts to loosen the above restrictions, most of which focus on dealing with the photometric variation. Those techniques includes: estimating illumination parameters in addition to motion parameters[5, 36] ; using intrinsically robust similarity measures, such as Mutual Information [15, 10], or normalized correlation [18]; preprocessing images [26, 3, 38], or using hand-crafted features such as HOG [9], SIFT [24], and LBP [31, 2] to obtain a representation that is more robust to intensity variations. However, none of them is robust against all these variations or shows competitive results in object tracking.

Following the recent trend of learning deep feature representation, in this work we propose to improve the robustness of LK by learning a feature representation specifically to the needs of LK framework. With the learned feature representation, our Deep-LK tracker is the first one in the LK family to achieve competitive results on challenging object tracking datasets, and significantly outperforms GOTURN, the state-of-the-art deep regression-based tracker.

## 3. Deep-LK Regressor

The validity of LK algorithm is based on the consistency of the linear relationship between the pixel intensity/feature value and warp parameter. Thus to improve the robustness of LK against the aforementioned variations in videos, the idea is to learn a deep convolution network that projects the template and source images to a feature space where this linear relationship is best approximated.

Deep convolutional features pre-trained on large-scale datasets (*e.g.* ImageNet [33]) have shown great generalizability to a variety of tasks, and demonstrate the ability to capture high-level semantics robust to variations such as scale, illumination, occlusions, *etc*. However, vanilla convolutional features trained for image classification are not optimized for LK. By integrating LK into the end-to-end deep learning framework, the cost curves of learned features become smoother with a single local minimum at the identity displacement (Fig. 2). These cost curves have a near-quadratic shape, which is ideal for the Newton's update of LK.

### 3.1. Network Architecture

Fig. 3 summarizes the network architecture of Deep-LK. Given an input source image $\mathcal{I}$, we extract deep convolutional features $\phi(\cdot)$ and apply a single linear regression layer in the feature space to predict the warp update. The linear regression layer is formulated as:

$$\Delta\tilde{\mathbf{p}} = \mathbf{R}_{\phi(\mathcal{T})} \cdot \phi(\mathcal{I}(\mathbf{p})) + \mathbf{b}_{\phi(\mathcal{T})}. \quad (11)$$

Here, to simplify notation, we abbreviate $\mathcal{T}(\mathbf{0})$ as $\mathcal{T}$.

Similar to IC-LK, $\mathbf{R}_\phi$ and $\mathbf{b}_\phi$ are formed through

$$\mathbf{R}_{\phi(\mathcal{T})} = \mathbf{W}^\dagger_{\phi(\mathcal{T})} \quad (12)$$
$$\mathbf{b}_{\phi(\mathcal{T})} = -\mathbf{R}_{\phi(\mathcal{T})} \cdot \phi(\mathcal{T}) \; , \quad (13)$$

where $\mathbf{W}_{\phi(\mathcal{T})} = \frac{\partial\phi(\mathcal{T})}{\partial\mathbf{p}}$ is computed in the same way as IC-LK. In contrast to conventional linear layers that are learnable itself, our regression matrix and bias pair $(\mathbf{R}_{\phi(\mathcal{T})}, \mathbf{b}_{\phi(\mathcal{T})})$ is formed following the IC-LK algorithm, and as a result, it is dependent on the template image $\mathcal{T}$ and the feature extraction function that follows; on the other hand, unlike IC-LK, the adopted feature $\phi(\cdot)$ is *learnable* instead of being hand-crafted like SIFT or HOG.

### 3.2. The Conditional LK Loss

There are several ways of formulating the learning objective to enforce the consistency of the linear relationship between feature values and warp parameters. One option is to use a *generative loss*, which directly penalizes the difference between the source image and the linear approximation of the template image with a ground truth warp parameter. Another option is to train the feature representation such that the warp parameter prediction is close to the ground truth. We refer to this objective as the *conditional LK loss*.

It has been shown in [23] that optimizing in terms of a conditional objective usually leads to better prediction than a generative objective. In our primary experiments, we also confirm that in our case, using conditional LK loss leads to far better tracking accuracy. This finding is in line with the benefits of an end-to-end training paradigm in the deep learning community.

Formally, the conditional LK loss is defined as:

$$\sum_{n\in\mathcal{S}} \mathcal{L}\left(\mathbf{R}_{\phi(\mathcal{T}^{(n)})} \cdot \phi(\mathcal{I}^{(n)}(\mathbf{p})) + \mathbf{b}_{\phi(\mathcal{T}^{(n)})} - \Delta\mathbf{p}_{gt}^{(n)}\right), \quad (14)$$

where $\mathcal{S}$ is the index set of all training data, $\mathcal{L}$ is the loss function, and $\mathbf{p}_{gt}$ is the ground truth warp parameter. This is optimized over the parameters of the feature extraction function $\phi(\cdot)$. We choose to use the Huber loss for optimization; compared to SSD, Huber loss is less sensitive to outliers in the data.

The conditional LK loss is differentiable and can be incorporated into a deep learning framework trained by backpropagation. Please refer to the supplementary materials for a detailed mathematical derivation.

### 3.3. Training/Online Adaptatinon

**Training:** We train Deep-LK with pairs of source and template images sampled from video sequences. First, a template image is generated for each frame by cropping an image region from an object bounding box. For each template image, we randomly sample a set of source images from the next frame. The source images are sampled by randomly perturbing the bounding box to mimic small motion changes between frames.

We follow the practice in GOTURN, which assumes the motion between frames follows a Laplace distribution. The scale changes are modeled as a Laplace distribution with zero mean and scale $b_s = 1/30$; The translations of bounding box are sampled using $b_x = 0.06$. In addition, we further augment the training set by adding random brightness, contrast and saturation shift to both the template and source images.

Unlike GOTURN, we do not use synthetically perturbed images from ImageNet to augment the training set. While all the assumptions for LK such as photometric consistency and the model of geometric deformation hold in such ideal scenarios offers little help for improving the feature robustness in real video sequences, where appearance/geometric variations can be more complex.

**Online adaptation:**

When tested on video, we follow the common practice in correlation filter trackers, which uses a simple template adaptation strategy:

$$\hat{\phi_{\mathcal{T}}}^{t} = (1-\alpha) \cdot \hat{\phi_{\mathcal{T}}}^{t-1} + \alpha \cdot \phi(\mathcal{I}(\mathbf{p}^{t-1})), \quad (15)$$

where $\hat{\phi_{\mathcal{T}}}^{t}$ represents the template feature used to generate linear regression parameters at frame $t$, and $\alpha$ is the learning rate. The computation cost of this online adaptation strategy is cheap, and it is effective to improve the robustness to pose, scale and illumination changes.

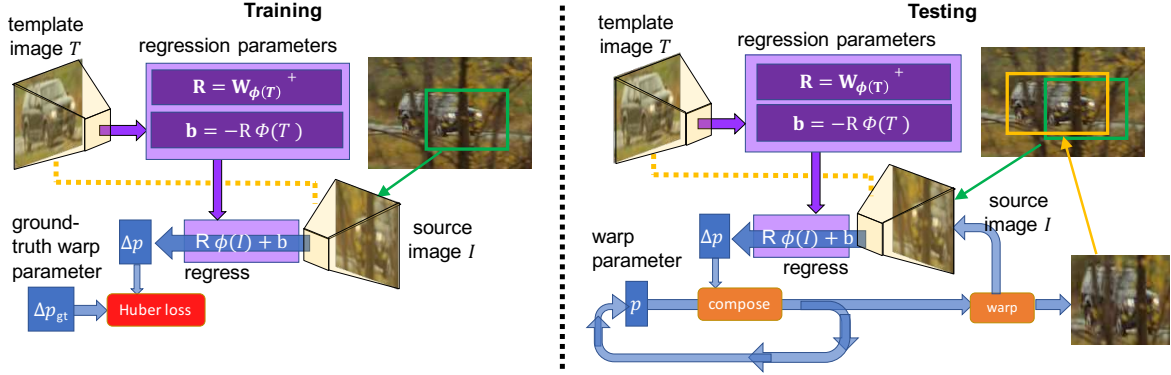

Figure 3. (left) We train the feature representation $\phi$ such that the warp parameter prediction is close to the ground truth; (right) In test time, given a pair of template image and a source image, Deep-LK first computes the feature for the template image, and forms the regression matrix $\mathbf{R}$ and bias term $\mathbf{b}$. Then in each iteration, Deep-LK computes features for the current source image, estimates the warp by regression, warps the source image and proceeds to the next iteration or stops when the box change of the source image is negligible.

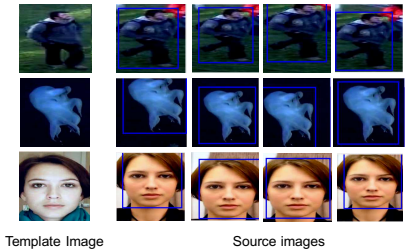

Figure 4. Source and template image pairs for training.

## 4. Experiment

### 4.1. Implementation Details

In this work, we use the conv5 feature of AlexNet [21] as the deep convolutional feature extraction function $\phi$, which maps an RGB image of size $227 \times 227$ to a feature map of size $13 \times 13$ with 256 filter channels.

We collect our training samples from ALOV300++ [34], which consists of 307 videos (7 videos are removed because of overlapping with VOT2014). Limited by the number of videos available, we only finetune the last convolution layer to avoid over-fitting.

Although Deep-LK is trained to be able to deal with both translation and scale estimation, we find that first updating translation with scale fixed until convergence, then updating translation and scale together works best in practice.

In addition, to avoid tracking failure caused by unreliable scale estimation, we implement an early stopping mechanism in the LK update. At each iteration, we check if the square distance between the source image and template image is decreasing, otherwise, we stop the update and report the warp parameter corresponding to the lowest square distance.

For 30 Frames Per Second (FPS) videos, we empirically set $\alpha = 0.03$, and for 240 FPS videos, the learning rate is decreased proportion to the change of frame rate, which is 0.0037.

### 4.2. Experiment Setup

We evaluated the proposed Deep-LK tracker over three sets of sequences including the VOT14 dataset [29], a set of 50 higher frame rate videos borrowed from the NfS dataset [1], and a small set of 15 videos captured by an higher frame rate (240 FPS) Unmanned aerial vehicle (UAV) camera from ground objects.

The sequences from the NfS dataset display generic objects in the real-world scenarios captured by 240 FPS smart phone cameras (*e.g.* iPhone6). The UAV videos allow us to evaluate the generalization capacity of our tracker on unseen viewpoints and objects. The reason we tested the Deep-LK on videos with different capture rates (30 FPS in VOT14, 240 FPS in NfS and UAV videos) is to explore the robustness of our method for both lower and higher frame rate videos. We compared our proposed algorithm against several state of the art object trackers including deep trackers (SiameseFc [7], GOTURN [16], MDNet [30] and FCNT [37]), CF trackers with hand-crafted features (SRDCF [13], Staple [6], DSST [12], SAMF [22], LCT [28], KCF [17] and CFLB [19]), and CF trackers with deep features (HCF [27] and HDT [32]). We do not report results on other public datasets, like VOT15 [20] and OTB [39] because our training set from ALOV++ is overlapped with their test videos.

**Evaluation Methodology:** We use the success metric to evaluate all trackers [39]. Success measures the intersection over union (IoU) of predicted and ground truth bounding boxes. The success plot shows the percentage of bounding boxes whose IoU is larger than a given threshold. We use the Area Under the Curve (AUC) of success plots to rank the trackers. We also compare all the trackers by the success rate at the conventional threshold of 0.50 (IoU > 0.50) [39].

### 4.3. Feature Evaluation

The first experiment evaluates the effect of learned features on Deep-LK's tracking accuracy. To do so, we run the Deep-Lk tracker using two different sets of deep features, including features we learn over the conditional LK loss, and those which directly borrowed from AlexNet (trained on ImageNet). We also explore the effect of utilizing features from different layers (*e.g.* Conv3, Conv4 and Conv5). The result on VOT14 sequences is presented in Fig. 5, showing that first, compared to AlexNet features, those features leaned over the conditional LK loss offer much higher accuracy and robustness. Second, Conv5 outperformed Conv3 and Conv4. We also provided attribute based comparison of such features in Fig. 6. Similarly, this evaluation shows that for all attributes features learned by the conditional LK outperform those from AlexNet. Moreover, among different feature layers learned by the conditional LK loss, Conv5 features- in general- are more robust on challenging situations such as camera motion, illumination change and occlusion. Based on this, we choose to use the Conv5 learned features in our Deep-LK framework.

### 4.4. Evaluation on VOT14

Fig. 7 illustrates the evaluation of Deep-LK on 25 challenging videos of the VOT14 dataset. Following the protocol in Fig. 7, results are reported by the means of accuracy and robustness [1]. The comparison shows the superior accuracy and robustness of our method to the leading methods in the VOT14 challenge, such as DSST, KCF, SAMF and GO-TURN. More specifically, our method outperformed the CF tracker KCF and deep tracker GOTURN. This shows the advantage of our method from two different perspectives. First, unlike KCF which only relies on online adaptation of hand-crafted features, Deep-LK adapts the regressor using discriminative deep features. This offers a well-generalized tracker which is more robust against challenging circumstances such as deformation and scaling. Second, compared to GOTURN- which is the only compared deep regression based tracker, Deep-LK delivers much more robust tracking, due to its crucial ability to update the regressor with new frames. This not only improves the generalization of this tracker to unseen objects, but increases its robustness to geometric and photometric variations between two consecutive frames. Compared to Deep-LK, MDNet achieved superior robustness and accuracy rank. The best overall robustness and accuracy is achieved by MDNet. This tracker, however, suffers from very expensive computation, such that its tracking speed on a high-end GPU is limited to ∼3 FPS, while Deep-LK on lower frame rate sequences of VOT14 can track up to 75 FPS.

[1]Readers are referred to Fig. 7 for more details.

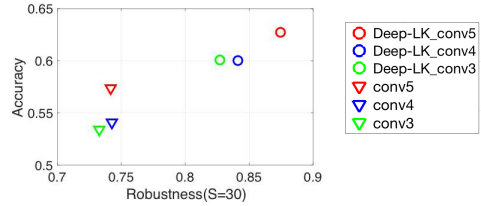

Figure 5. Accuracy and robustness plot on VOT14 dataset, evaluating Deep-LK tracking performance using different layers of learned and AlexNet's features.

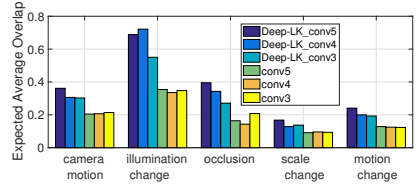

Figure 6. Comparing the effect of AlexNet features with those learned over the conditional LK loss on tracking performance of Deep-LK. Results are reported by the means of average overlap for Conv3, Conv4 and Conv5 layers, on 5 attributes of VOT14.

### 4.5. Evaluation on NfS Sequences

The results of this experiment are shown in Table 1 and Fig. 8, comparing Deep-LK with the state of the art CF and deep trackers on 50 NfS videos. These sequences are captured by 240 FPS cameras and, thus, display less visual changes between two consecutive frames [1]. This is desirable characteristic for Deep-LK which directly regresses from current frame to the next one with much less appearance changes. As summarized in Table 1, our method achieves the highest AUC (51.2), closely followed by MD-Net (50.9) and SRDCF (50.7). This result shows that on higher frame rate videos, our regression based tracker can perform as accurate as such classification/detection based methods. Our method also demonstrates a significant improvement to GOTURN (∼10 %). This is mainly because of Deep-LK's ability to adapt its regressor to the appearance of the currently tracker frame. This crucial adaptation capacity which is missed in GOTURN can easily increase its sensitivity to unseen scenarios.

In terms of tracking speed on the CPU, our method is faster than all deep trackers as well as several CF trackers (SRDCF, LCT, DSST, HCF and HDT). On the GPU, however, GOTURN showed a tracking speed of 155 FPS which is 50% faster than Deep-LK on the same GPU. During this experiment, we observed that Deep-LK performs much faster on higher frame rate videos. The reason is that since inter-frame difference in higher frame rate videos is much less than lower frame rate ones, Deep-LK requires less iterations to converge. This offers a tracking speed of 100 FPS on GPUs which is 25% faster than tracking lower frame rate sequences of VOT14 (75 FPS). Fig. 10 (a) visualizes tracking performance of Deep-LK with SRDCF, Staple, MDNet and GOTURN on three different scenarios. This qualitative result demonstrates the robustness of our method against severe scale changes, out-plane-rotation, occlusion and non-rigid deformation.

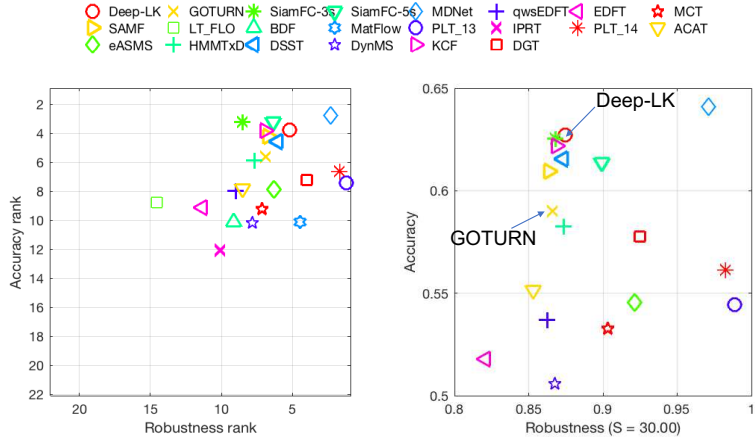

Figure 7. Results on VOT14, comparing Deep-LK with the leading methods in the VOT14 challenge. Best trackers are closer to the top right corner.

| | Deep-LK | SRDCF | Staple | LCT | DSST | SAMF | KCF | CFLB | HCF | HDT | MDNet | SFC | FCNT | GOTURN |
|---|---|---|---|---|---|---|---|---|---|---|---|---|---|---|
| AUC | 51.2 | 50.7 | 45.4 | 37.9 | 48.1 | 47.8 | 36.9 | 22.5 | 41.3 | 50.4 | 50.9 | 49.2 | 50.0 | 40.8 |
| Speed (CPU) | 20.7 | 3.8 | 50.8 | 10 | 12.5 | 16.6 | 170.4 | 85.5 | 10.8 | 9.7 | 0.7 | 2.5 | 3.2 | 3.9 |
| Speed (GPU) | 100 | - | - | - | - | - | - | - | - | 43.1 | 2.6 | 48.2 | 51.8 | 155.3 |

Table 1. Comparing Deep-LK with the recent CF trackers on 50 higher frame rate sequences- from NfS. Results are reported as AUCs of the success plot. Tracking speed is reported in Frames Per Second (FPS) on the CPU and/or GPU if applicable.

### 4.6. Robustness to Unseen Objects and Viewpoints

The goal of this experiment is to evaluate the robustness of Deep-LK to unseen objects and viewpoints, compared to GOTURN (as the baseline). For this purpose, we captured 15 challenging higher frame rate videos (240 FPS) using an UAV (drone) camera. These videos contain ∼100K frames showing real-world scenarios of different objects such as drone, car, truck, boat and human (biking, running, walking and playing basketball). The result in Fig. 9 demonstrates the notable accuracy of our method (60.38) compared to GOTURN (48.13). As mentioned earlier, the robustness of Deep-LK comes from its ability to adapt its regressor to the appearance of current frame. This, however, is not the case in GOTURN. Its regression function is frozen and, thus, performs poor on unseen targets and viewpoints. Fig. 10 (b) visualizes the poor generalization of GOTURN to the unseen aerial viewpoint (of a truck) and object (drone). It also shows that Deep-LK generalizes well to unseen viewpoint/object with challenging appearance variation.

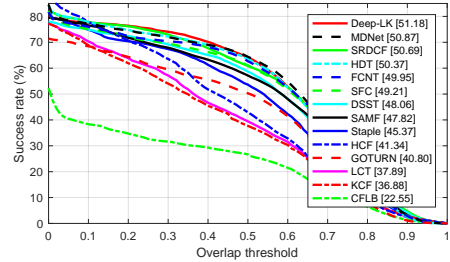

Figure 8. Comparing Deep-LK with the state of the art CF and deep trackers on 50 NfS videos. AUCs are reported in the legend.

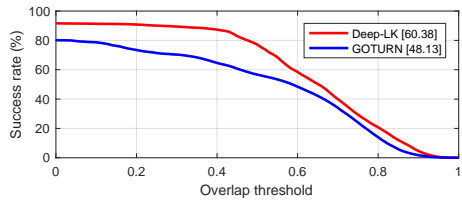

Figure 9. Comparison Deep-LK with GOTURN on 15 UAV sequences. AUCs are reported in the legend.

### 4.7. Speed Analysis of Deep-LK

As detailed in Section 3.4, Deep-LK does iterative update and stops when either it converges or meets other stopping criterion. Thus, its tracking speed is highly dependent on 1) the capture frame rate of the video, and 2) the motion of the target object. To analyze the speed of Deep-LK, we break down the computational cost of each operation within the Deep-LK framework. On GPU, for each frame, it takes about 1.4ms to form the regression parameters from template, and approximately 3ms per LK iteration (0.5ms for image cropping and resizing, $2.2 \sim 2.5$ms to compute Conv5 features and around 0.15ms to perform regression). On VOT14 with lower frame rate sequences, Deep-LK on average converges within 4 iterations per frame, hence our average speed is about 75 FPS on GPU. On NfS (and UAV) higher frame rate sequences, since there is much less inter-frame appearance changes, Deep-LK converges much faster over $2 \sim 3$ iterations per frame. This offers a tracking speed of 100 FPS on the same GPU. On CPU, on the other hand, Deep-LK requires relatively longer time to compute Conv5 features and update the regressor. As a result, it performs $\sim$15 FPS on lower frame rate videos of VOT14, and $\sim$20 FPS on higher frame rate videos of NfS and UAV data.

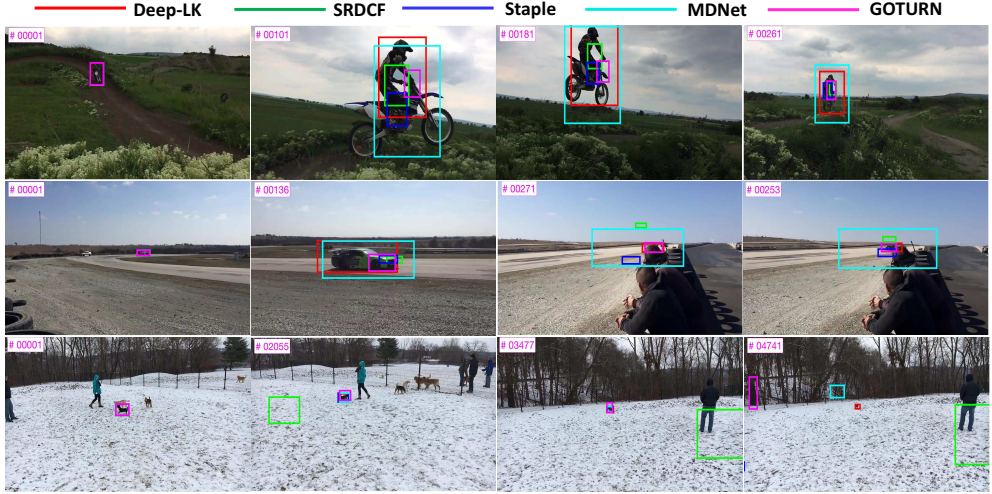

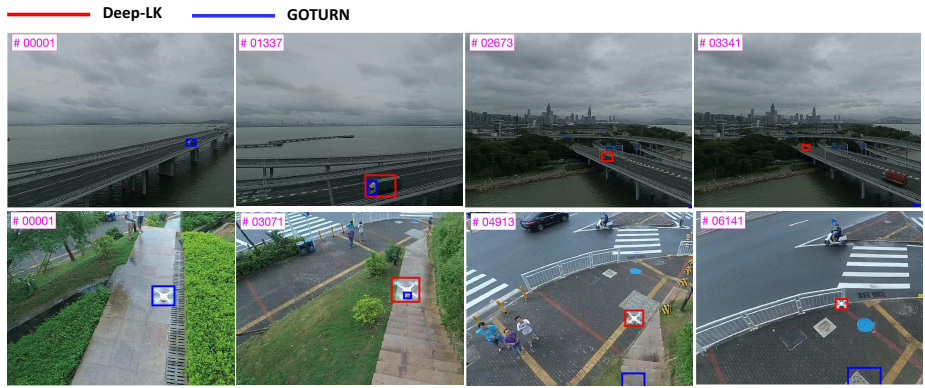

Figure 10. Qualitative results. (a) shows tracking performance of Deep-LK, SRDCF, Staple, MDNet and GOTURN on 4 higher frame rate videos selected from NfS. Our method is robust against scaling, non-rigid deformation, out-of-plane rotation and occlusion. (b) comparing the robustness of our method to unseen object and viewpoint compared to GOTURN.

## 5. Conclusion

We propose a novel regression-based object tracking framework, which successfully incorporates Lucas & Kanade algorithm into an end-to-end deep learning paradigm. We conclude that the combination of offline trained feature representation and the efficient online adaptation of regression parameters respect to template images, are crucial advantages of our method to generalize well against real-world siturations such as severe scaling, deformation and unseen objects/viewpoints. Compared to GOTURN- the state of the art regression based deep tracker- that only exploits offline trained regression parameters, our Deep-LK shows impressive robustness to unseen objects and viewpoints. Moreover, we demonstrate that on higher frame rate sequences, Deep-LK offers comparable tracking performance to current state of the art trackers including both correlation filter and deep framework tracker.

# Appendix: Differentiation of the Conditional LK Loss

Here, we derive the differentiation of the Conditional LK loss for back-propagation. First, for convenience, we repeat the definition of conditional LK loss:

$$\min_{\theta} \sum_{n \in \mathcal{S}} \mathcal{L}\left(\mathbf{R}_{\phi(\mathcal{T}^{(n)})} \cdot \phi(\mathcal{I}^{(n)}(\mathbf{p})) + \mathbf{b}_{\phi(\mathcal{T}^{(n)})} - \Delta \mathbf{p}_{gt}^{(n)}\right), \quad (16)$$

where $\mathcal{L}$ is defined as Huber loss, $\phi(\cdot)$ refers to the feature extraction function, and $\theta$ is the model parameter of $\phi$.

By noticing the fact that:

$$\mathbf{b}_{\phi(\mathcal{T})} = -\mathbf{R}_{\phi(\mathcal{T})} \cdot \phi(\mathcal{T}), \quad (17)$$

the conditional LK loss can be rewritten as:

$$\sum_{n \in \mathcal{S}} \mathcal{L}\left(\mathbf{R}_{\phi(\mathcal{T}^{(n)})} \cdot \left[\phi(\mathcal{I}^{(n)}(\mathbf{p})) - \phi(\mathcal{T}^{(n)})\right] - \Delta \mathbf{p}_{gt}^{(n)}\right). \quad (18)$$

Hence, the partial derivative of the conditional LK loss over the parameter of feature extraction function $\phi$ can be computed as:

$$\frac{\partial \mathcal{L}}{\partial \theta_i} = \frac{\partial \mathcal{L}}{\partial \phi(\mathcal{T})} \frac{\partial \phi(\mathcal{T})}{\partial \theta_i} + \frac{\partial \mathcal{L}}{\partial \phi(\mathcal{I}(\mathbf{p}))} \frac{\partial \phi(\mathcal{I}(\mathbf{p}))}{\partial \theta_i}, \quad (19)$$

where,

$$\frac{\partial \mathcal{L}}{\partial \phi_i(\mathcal{T})} = \mathcal{F}\left(\left(\mathbf{R}_{\phi(\mathcal{T})} \cdot [\phi(\mathcal{I}(\mathbf{p})) - \phi(\mathcal{T})] - \Delta \mathbf{p}_{gt}\right)^T\right)$$
$$\cdot \left(\frac{\partial \mathbf{R}_{\phi(\mathcal{T})}}{\partial \phi_i(\mathcal{T})} [\phi(\mathcal{I}(\mathbf{p})) - \phi(\mathcal{T})] - \mathbf{R}_{\phi(\mathcal{T})} \delta_i\right), \quad (20)$$

and

$$\frac{\partial \mathcal{L}}{\partial \phi_i(\mathcal{I}(\mathbf{p}))} = \mathcal{F}\left(\left(\mathbf{R}_{\phi(\mathcal{T})} \cdot [\phi(\mathcal{I}(\mathbf{p})) - \phi(\mathcal{T})] - \Delta \mathbf{p}_{gt}\right)^T\right)$$
$$\cdot \mathbf{R}_{\phi(\mathcal{T})} \delta_i. \quad (21)$$

Here, $\delta_i$ denotes a one-hot vector with $1$ at dimension $i$, and $\mathcal{F}$ represents the derivative of the Huber loss, which is a truncation function clipping the value of each dimension of the input vector:

$$\mathcal{F}(\mathbf{x}) = \begin{bmatrix} \vdots \\ \min(\max(x_i, -1), 1) \\ \vdots \end{bmatrix}.$$

So far, there lefts one last missing piece to complete the differentiation of the conditional LK loss - the partial derivative of $\mathbf{R}_{\phi(\mathcal{T})}$ over $\phi_i(\mathcal{T})$ in Equation 20.

Recall that $\mathbf{R}_{\phi(\mathcal{T})}$ is the pseudo inverse of $\mathbf{W}_{\phi(\mathcal{T})}$ (abbreviated as $\mathbf{W}$ for conciseness), we have:

$$\frac{\partial \mathbf{R}_{\phi(\mathcal{T})}}{\partial \phi_i(\mathcal{T})} = \frac{\partial (\mathbf{W}^T\mathbf{W})^{-1}}{\partial \phi_i(\mathcal{T})}\mathbf{W}^T + (\mathbf{W}^T\mathbf{W})^{-1}\frac{\partial \mathbf{W}^T}{\partial \phi_i(\mathcal{T})}, \tag{22}$$

where,

$$\frac{\partial (\mathbf{W}^T\mathbf{W})^{-1}}{\partial \phi_i(\mathcal{T})} = -(\mathbf{W}^T\mathbf{W})^{-1}\bigg(\mathbf{W}^T\frac{\partial \mathbf{W}}{\partial \phi_i(\mathcal{T})} + \frac{\partial \mathbf{W}^T}{\partial \phi_i(\mathcal{T})}\mathbf{W}\bigg)(\mathbf{W}^T\mathbf{W})^{-1}, \tag{23}$$

and

$$\frac{\partial \mathbf{W}}{\partial \phi_i(\mathcal{T})} = \begin{bmatrix} \ddots & & \\ & \frac{\partial \nabla \phi_j(\mathcal{T})}{\partial \phi_i(\mathcal{T})} & \\ & & \ddots \end{bmatrix} \begin{bmatrix} \vdots \\ \frac{\partial \mathcal{W}(x_j,\mathbf{0})}{\partial \mathbf{p}} \\ \vdots \end{bmatrix}. \tag{24}$$

Here, $\nabla \phi_j(\mathcal{T})$ is estimated by finite-differencing, therefore, Equation 24 is computed analytically.

Now that we have obtained the expression of $\frac{\partial \mathcal{L}}{\partial \phi_i(\mathcal{T})}$ and $\frac{\partial \mathcal{L}}{\partial \phi_i(\mathcal{I}(\mathbf{p}))}$ (Equation 20 and 21), we can thus run back-propagation (Equation 19) to compute $\{\frac{\partial \mathcal{L}}{\partial \theta_i}\}$, which is the gradient of feature extraction parameters.